\documentclass{llncs}

\usepackage{times}
\usepackage{url}
\usepackage{booktabs}
\usepackage{graphicx}

\usepackage[sort]{cite}

\begin{document}

\title{Ranking Algorithms by Performance}
\author{Lars Kotthoff}
\institute{INSIGHT Centre for Data Analytics}

\maketitle

\begin{abstract}
A common way of doing algorithm selection is to train a machine learning model
and predict the best algorithm from a portfolio to solve a particular problem.
While this method has been highly successful, choosing only a single algorithm
has inherent limitations -- if the choice was bad, no remedial action can be
taken and parallelism cannot be exploited, to name but a few problems. In this
paper, we investigate how to predict the ranking of the portfolio algorithms on
a particular problem. This information can be used to choose the single best
algorithm, but also to allocate resources to the algorithms according to their
rank. We evaluate a range of approaches to predict the ranking of a set of
algorithms on a problem. We furthermore introduce a framework for categorizing
ranking predictions that allows to judge the expressiveness of the predictive
output. Our experimental evaluation demonstrates on a range of data sets from
the literature that it is beneficial to consider the relationship between
algorithms when predicting rankings. We furthermore show that relatively na\"ive
approaches deliver rankings of good quality already.
\end{abstract}

\section{Introduction}

The Algorithm Selection Problem \cite{rice_algorithm_1976} is to select the most
appropriate algorithm for solving a particular problem. It is especially
relevant in the context of algorithm portfolios
\cite{huberman_economics_1997,gomes_algorithm_2001}, where a single solver is
replaced with a set of solvers and a mechanism for selecting a subset to use on
a particular problem.

The most common approach to building such a selector is to use machine learning
to induce a model of the algorithms in the portfolio. Such models can take many
different shapes and forms. Common ones include classification or clustering
approaches that select only a single solver to be run and regression approaches
that predict the performance of each portfolio solver independently and choose
one or several according to those predictions. While some of these approaches
implicitly compute a ranking of the algorithms in the portfolio, almost none
make explicit use of it.

In this paper, we investigate approaches that explicitly rank the portfolio
algorithms according to their performance. Instead of predicting only the best
algorithm, we compute a ranking over \emph{all} of them. There are several
advantages of this approach. Because we have information on all algorithms, we
can use several or all of them for solving the problem or set of problems. If
the algorithm that is chosen initially performs worse than expected, the next
best can be chosen without having to make another prediction. If more than one
processor is available, several can be run in parallel. This again limits the
impact of bad predictions.

Several approaches in the literature, e.g.\
\cite{omahony_using_2008,kadioglu_algorithm_2011}, compute schedules for
running the algorithms in the portfolio. Such schedules rely on a ranking of the
algorithms that dictates when to run each algorithm and for how long. Despite
this, no comparison of different ways of arriving at such a schedule has been
performed to date.

While techniques that could also be used to derive a performance ranking of
algorithms are abundant in the literature, very few researchers have
investigated how and to what extent performance rankings can be predicted in
practice. \cite{kotthoff_hybrid_2012} shows that the prediction of the
performance of target algorithms makes incorrect predictions very often and
proposes a means of mitigating this, but again only for the best algorithm. This
suggests that more sophisticated means for predicting complete rankings may be
required.

In this paper, we propose and evaluate a range of different approaches,
including ones that are commonly used in the literature for similar purposes, on
data sets that we also take from the literature. Our evaluation fills a
much-needed gap by addressing the issue of how one should predict complete
performance rankings for algorithms in a portfolio context.

While a complete ranking is not required to do algorithm selection, it can be
beneficial. Predictions of algorithm performance will always have some degree of
uncertainty associated with them and being able to choose from among all
portfolio algorithms in a meaningful way can be used to mitigate the effect of
this. However, in order for this to be exploited, we must be able to predict the
rankings to be used reliably and know what techniques are available for doing
so. This is the core contribution of this paper.

We furthermore propose a framework for the categorization of types of
predictions from which rankings can be derived. This framework serves as a means
of clarifying and formalising the problem. In particular, it allows researchers
to access the type of ranking predictions they are making with respect to how
fine-grained the decisions to be taken based upon the ranking can be.

\section{Background}

Predicting a ranking of entities is not a new problem. In machine learning, the
problem is known as label ranking and a large body of literature related to it
exists. A review of this literature is beyond the scope of this paper; the
interested reader is referred to a recent survey in \cite{furnkranz_label_2011}.
In this paper, we focus on the problem of ranking in the algorithm selection
context. Therefore, the approaches investigated here are based on existing
techniques for algorithm selection rather than label ranking. While there are
many other approaches, we believe that the ones presented here make most sense
in this context and will be most familiar to algorithm selection researchers.

There exist many approaches to solving the Algorithm Selection Problem by
different means in various contexts and even a cursory survey is beyond the
scope of this paper. The interested reader is referred to a recent survey
presented in \cite{kotthoff_algorithm_2012}. We focus on the approaches that are
immediately relevant to this paper.

\cite{brazdil_ranking_2003} propose and evaluate a framework to generate
rankings of machine learning algorithms. Their approach predicts a compound
measure that takes accuracy and time into account and derives the ranking
directly from this. In later work \cite{soares_meta-learning_2004}, they
propose and evaluate a method to retrieve and combine rankings on training data.
This work considers the algorithms in the portfolio in isolation and derives
rankings from the combination of the independent predictions.

More recently, \cite{kanda_meta-learning_2012} proposed an approach for ranking
algorithms that explicitly uses label ranking techniques from machine learning.
The authors use a neural network with an output neuron for each algorithm in the
portfolio that identifies its rank.

In contrast to this, \cite{hurley_adaptation_2012} use a
voting-based approach to derive the ranking of a set of algorithms. They use a
$k$-nearest neighbour model to identify the cases that are most similar to the
one to solve. Instead of predicting a score or a ranking, they count the best
algorithm for each retrieved case as a vote and rank the algorithms according to
that.

\cite{leite_active_2010} derive a ranking by making pairwise comparisons between
algorithms and focus on reducing the number of comparisons that have to be made
by using active learning techniques. The ranking is based on the partial order
computed using the set of comparisons. \cite{kotthoff_evaluation_2012} use
statistical relational learning to predict the complete ranking directly of a
set of algorithms on a problem. They do however select only the single best
algorithm and report that this particular approach is not competitive.

A number of algorithm selection approaches compute schedules for running the
algorithms in a portfolio. Some of them do this implicitly by allocating
resources to the algorithms, some determine explicit schedules. In both cases,
rankings are used implicitly or explicitly to arrive at the schedule.
\cite{omahony_using_2008} find the examples that are closest to the problem to
solve in a case base and compute a schedule of solvers to run based on the
performance on those problems. They do not predict a ranking, but compute it
based on the observed performance.

\cite{kadioglu_algorithm_2011} compute schedules of algorithms to run in a
similar way, but focus on the problem of computing the schedule once the ranking
is known. They also determine the ranking based on the observed performance on a
set of training problems. Other approaches that compute (implicit) rankings for
the purpose of determining when and for how long to run algorithms include
\cite{howe_exploiting_1999,roberts_directing_2006,streeter_combining_2007,pulina_self-adaptive_2009}.

\section{Organizing predictions}

Machine learning models are capable of making different types of predictions,
depending on the leaner that was used to induce the model. Classification models
predict labels, whereas regression models predict numbers. Statistical
relational learning is a relatively new area of machine learning that allows to
make complex predictions.

The different types of predictions can then in turn be used in different ways to
obtain rankings. We propose the following levels to categorise the predictive
output of a model with respect to what ranking may be obtained from it. The
model was heavily inspired by the theory of scales of measurement
\cite{stevens_theory_1946}. The levels proposed here correspond to the types of
scales.

\begin{description}
\item[Level 0] The prediction output is a single label of the best algorithm.
    This is the approach standard classification takes. It is not possible to
    construct a ranking from this and we do not consider it in this paper.
    Level~0 corresponds to the nominal scale in the theory of scales of
    measurements.
\item[Level 1] The prediction output is a ranking of algorithms. This ranking
    may be derived from a set of intermediate predictions. The relative position
    of algorithms in the ranking gives no indication of the difference in
    performance. That is, the performance difference between the algorithms at
    rank one and two may be much higher than the difference between the
    algorithms at rank two and three, but there is no way of computing this.
    Statistical relational learning approaches are capable of making such
    predictions. Level~1 corresponds to the ordinal scale.
\item[Level 2] The prediction output is a ranking with associated scores. Again
    the ranking may be derived from intermediate predictions. An example of this
    approach would be to predict the performance of each algorithm individually
    and construct the ranking from the predictions. The difference between
    ranking scores is indicative of the difference in performance. Level~2
    corresponds to the interval scale.
\end{description}

The theory of scales of measurements proposes the ratio scale as an additional
level. This is not required in our framework as it does not add any information
that would be useful for the purpose of ranking algorithms in a portfolio. The
ratio scale would for example correspond to the ratio of the ranking scores to
a baseline, e.g.\ the performance of the model that always chooses the single
overall best algorithm.

In the remainder of this paper, we will denote the framework $\mathcal{R}$ and
level $x$ within it $\mathcal{R}_x$. Note that higher levels strictly dominate
the lower levels in the sense that their predictive output can be used to the
same ends as the predictive output at the lower levels.

In the context of algorithm selection and portfolios, examples for the different
levels are as follows. A $\mathcal{R}_0$ prediction is suitable for selecting a
single algorithm. $\mathcal{R}_1$ allows to select the $n$ best solvers for
running in parallel on an $n$ processor machine. $\mathcal{R}_2$ allows to
compute a schedule where each algorithm is allocated resources according to its
rank and score. Note that while it is possible to compute a schedule given just
a ranking with no associated scores (i.e.\ $\mathcal{R}_1$), a much more
fine-grained schedule can usually be computed with scores.

\section{Empirical evaluation}

The aim of the empirical investigation is to identify which approaches and
methods for predicting the rank of a set of algorithms on a problem are likely
to achieve good performance. We investigate the performance of a number of
different approaches on several data sets from the algorithm selection
literature.

\subsection{Rank prediction approaches}

We evaluate the following ten ways of ranking algorithms, five from
$\mathcal{R}_1$ and five from $\mathcal{R}_2$.

\begin{description}
\item[Order] The ranking of the algorithms is predicted directly as a label. The
    label consists of a concatenation of the ranks of the algorithms. This
    approach is in $\mathcal{R}_1$. \cite{kotthoff_evaluation_2012} use a
    conceptually similar approach to compute the ranking with a single
    prediction step.
\item[Order score] For each training example, the algorithms in the portfolio
    are ranked according to their performance. The rank of an algorithm is the
    quantity to predict. We used both regression and classification approaches.
    The ranking is derived directly from the predictions. These two approaches
    are in $\mathcal{R}_1$.
\item[Faster than classification] A classifier is trained to predict the
    ranking as a label similar to the approach above given the predictions of
    which is faster for each pair of algorithms. This approach is in
    $\mathcal{R}_1$.
\item[Faster than difference classification] A classifier is trained to predict
    the ranking as a label given the predictions for the performance
    differences for each pair of algorithms. This approach is in
    $\mathcal{R}_1$.
\item[Solve time] The time to solve a problem is predicted and the ranking
    derived directly from this. In addition to predicting the time itself, we
    also predicted the log. These approaches are in $\mathcal{R}_2$. Numerous
    approaches predict the solve time to identify the best algorithm, for
    example \cite{xu_satzilla_2008}.
\item[Probability of being best] The probability of being the best algorithm for
    a specific instance in a $[0,1]$ interval is predicted. The ranking is
    derived directly from this. This approach is in $\mathcal{R}_2$.
\item[Faster than majority vote] The algorithms are ranked by the number of
    times they were predicted to be faster than another algorithm. This is the
    approach used to identify the best algorithm in recent versions of
    SATzilla \cite{xu_hydra-mip_2011}. This approach is in $\mathcal{R}_2$.
    While the individual predictions are simple labels (faster or not), the
    aggregation is able to provide fine-grained scores.
\item[Faster than difference sum] The algorithms are ranked by the sum over the
    predicted performance differences for each pair of algorithms. Algorithms
    that are often or by a lot faster will have a higher sum and rank higher.
    This approach is in $\mathcal{R}_2$.
\end{description}

We do not evaluate any approaches based on statistical relational learning that
predict rankings directly without intermediate predictions.
\cite{kotthoff_evaluation_2012} report results that are not competitive using
such an approach. As statistical relation learning is a relatively new area of
machine learning, the number of available implementations is very limited and we
were unable to find a suitable approach.

\subsection{Data sets}

We evaluate the performance on four data sets taken from the literature. We take
two sets from the training data for SATzilla 2009. This data consists of SAT
instances from two categories -- hand-crafted and industrial. They
contain 1181 and 1183 instances and are denoted SAT-HAN and SAT-IND,
respectively. We use the same 91 attributes as the SATzilla authors to describe
each instance and select a SAT solver from a portfolio of 19 solvers for SAT-HAN
and 18 solvers for
SAT-IND\footnote{\url{http://www.cs.ubc.ca/labs/beta/Projects/SATzilla/}}. We
adjusted the timeout values reported in the training data available on the
website to 3600 seconds after consultation with the SATzilla team as some of the
reported timeout values are incorrect.

The third data set comes from the Quantified Boolean Formulae (QBF) Solver
Evaluation 2010\footnote{\url{http://www.qbflib.org/index_eval.php}} and
consists of 1368 QBF instances from the main, small hard, 2QBF and random
tracks. It is denoted QBF. 46 attributes are calculated for each instance and we
select from a portfolio of five QBF solvers. Each solver was run on each
instance for at most 3600 CPU seconds. If the solver ran out of memory or was
unable to solve an instance, we assumed the timeout value for the runtime. The
experiments were run on a machine with a dual four core Intel E5430 2.66\,GHz
processor and 16\,GB RAM. Our last data set, denoted CSP, is taken from
\cite{gent_learning_2010} and selects from a portfolio of two solvers for a
total of 2028 constraint problem instances from 46 problem classes with 17
attributes each.

\subsection{Methodology}

We use the Weka machine learning toolkit \cite{hall_weka_2009} to train models
and make predictions. As it is not obvious which machine learning algorithms
will perform well, we evaluated our approaches using the
\texttt{AdaBoostM1}
\texttt{BayesNet},
\texttt{DecisionTable},
\texttt{IBk} with 1, 3, 5 and 10 neighbours,
\texttt{J48},
\texttt{J48graft},
\texttt{JRip},
\texttt{LibSVM} with radial basis function kernel,
\texttt{MultilayerPerceptron},
\texttt{OneR},
\texttt{PART},
\texttt{RandomForest},
\texttt{RandomTree},
\texttt{REPTree},
and \texttt{SimpleLogistic}
algorithms for
classification and the
\texttt{AdditiveRegression},
\texttt{GaussianProcesses}
\texttt{LibSVM} with $\varepsilon$ and $\nu$ kernels,
\texttt{LinearRegression},
\texttt{M5P},
\texttt{M5Rules},
\texttt{REPTree},
and \texttt{SMOreg}
algorithms for regression. For all of these algorithms, we used the standard
parameters in Weka.

Our approaches are agnostic to the underlying machine learning algorithm that
is used to make the particular prediction that is required. We chose to include
a large number of different machine learning algorithms in our evaluation to be
able to judge objectively the performance of a particular approach rather than
the performance of the combination of an approach and an underlying machine
learning algorithm. By averaging over a number of machine learning algorithms,
we can mitigate any distorting effects individual combinations may have.

Our experiments include a total of 20 classification and 9 regression
algorithms. For some of the rank prediction approaches, it is necessary to use
several layers of machine learning algorithms. For example for the Faster than
difference classification approach, a regression algorithm is used to predict
the performance difference for each pair of algorithms and then a classification
algorithm to combine these predictions.

Where several layers of machine learning algorithms are required, they are
stacked as follows. The first layer is trained on the original training set with
the features of the original problems. The prediction of the models of this
first layer is used to train a model in a second layer that takes the
predictions of the earlier layer as input. The output is the final prediction
that we use to compute the ranking.

We evaluated all possible combinations of machine learning algorithms for all
approaches. That is, in the case where more than one layer of models is
required, we considered all combinations of algorithms for the several layers.
In total, 685 combinations were considered. In practice, some of these ran out
of memory or took too long to complete on some data sets.

The performance of each approach on each data set is evaluated using stratified
ten-fold cross-validation. The entire data set is partitioned into ten subsets
of roughly equal size with the roughly same distribution of best performing
algorithms. Nine of these ten sets are combined for training and the remaining
set is used for testing. The process is repeated with a different subset used
for testing until all subsets have been used for testing.

We assess the quality of a predicted ranking by comparing it to the actual
ranking using the Spearman correlation test. It returns a measure of association
between -1 and 1, indicating the predicted ranking is equal to the inverse of
the actual ranking or the actual ranking itself, respectively. We record the
quartiles of the association scores over all problem instances for a particular
data set, rank prediction approach and combination of machine learning
algorithms.

\section{Results}

We present summary results for the best and worst machine learning algorithms
for each prediction approach. These were determined as follows. For each
prediction type, the quartiles of the ranking scores for each machine learning
algorithm were summed over all data sets. The machine learning algorithms are
then ranked by this sum, where higher values are better. The algorithm with the
highest aggregate ranking score was the best, the one with the lowest score the
worst. Note that the best/worst algorithm for one approach was not necessarily
the best/worst for any of the other approaches.

\begin{figure*}[htb]
\centering
\includegraphics[width=\textwidth]{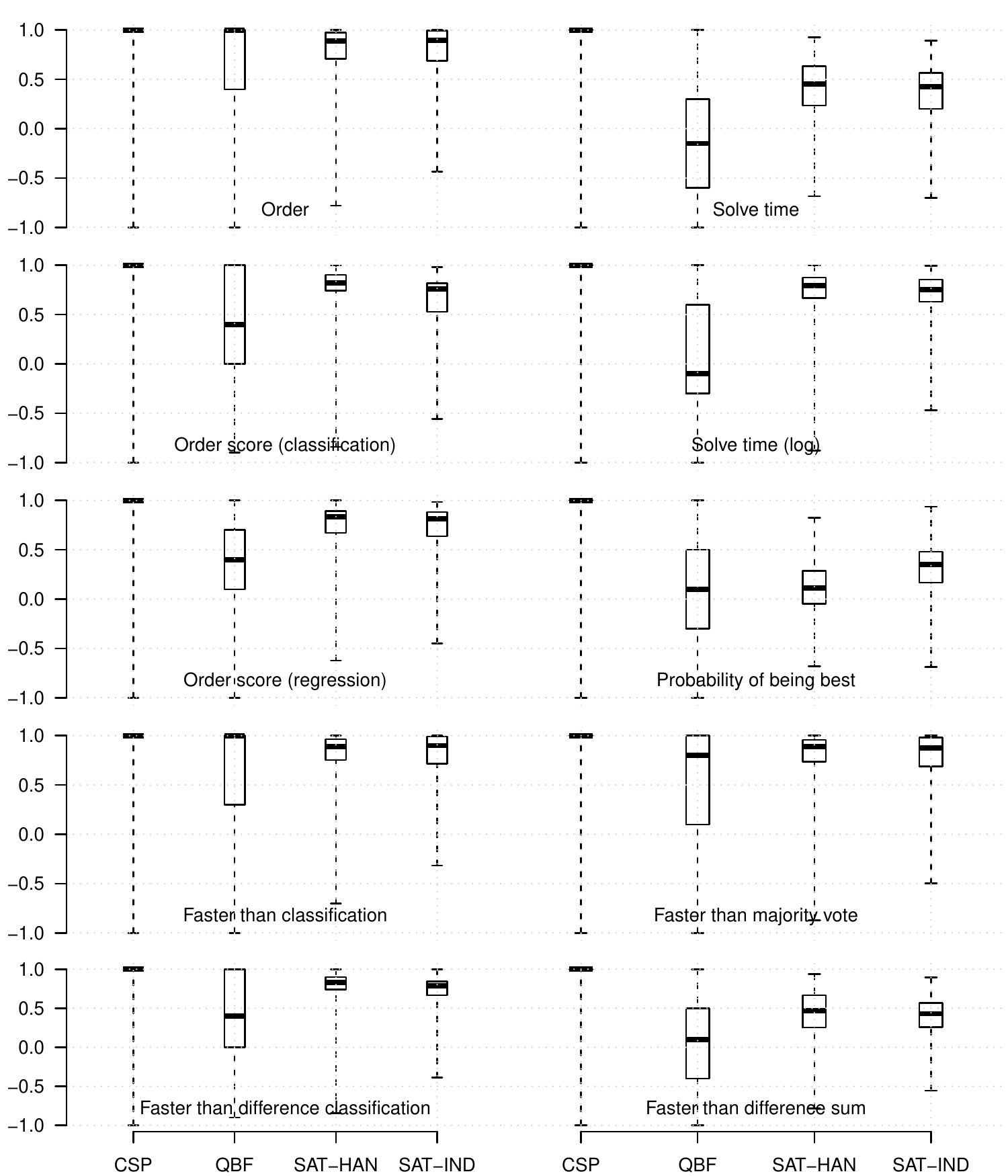}
\caption{Results for the best machine learning algorithm for each data set and
rank prediction approach. For each prediction approach, the best algorithm over
all data sets is shown. The thick line in the center of the box corresponds to
the median, the top and bottom ends of the box to the 75th and 25th percentile
respectively and the top and bottom end of the whiskers to the maximum and
minimum, respectively.\label{fig:resbest}}
\end{figure*}

\begin{figure*}[htb]
\centering
\includegraphics[width=\textwidth]{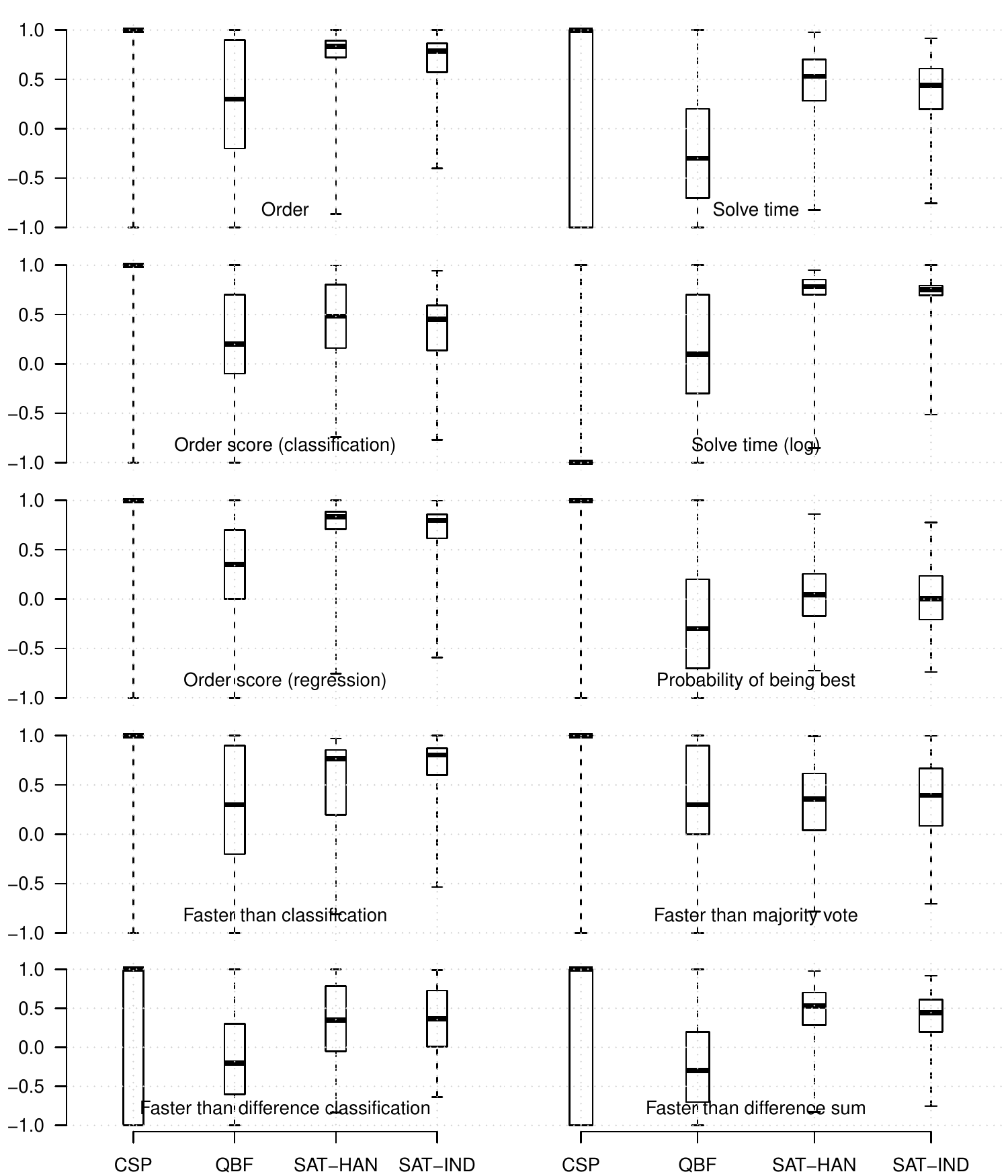}
\caption{Results for the worst machine learning algorithm for each data set and
rank prediction approach. For each prediction approach, the worst algorithm over
all data sets is shown. The thick line in the center of the box corresponds to
the median, the top and bottom ends of the box to the 75th and 25th percentile
respectively and the top and bottom end of the whiskers to the maximum and
minimum, respectively.\label{fig:resworst}}
\end{figure*}

Figures~\ref{fig:resbest} and~\ref{fig:resworst} give the results for best and
worst algorithms, respectively. The first observation is that for all rank
prediction approaches and data sets, there is a large spread in the quality of
the predicted rankings. However, most of the time the majority of predictions
for the best algorithm is closer to 1 than 0, meaning that good rankings are
achieved on average. In particular the median values are often very close to 1.

There are only two algorithms for the CSP data set and thus the ranking quality
score will always be 1 or -1. The worse score occurs only in a few cases even
for the worst machine learning algorithm and all approaches perform consistently
well. Unfortunately this does not give us any information as to which approach
is better. On the other data sets however, there is a larger number of
algorithms and thus a much richer set of possible rankings. The results achieved
on those data sets are more informative.

We are presenting the results for the both the best and the worst machine
learning algorithm for each rank prediction approach. This way, we can provide a
picture of the overall performance of each approach regardless of the underlying
machine learning algorithm. All the approaches presented in this paper are
agnostic to how the predictions they require are achieved.

In a real portfolio setting, the performance of the machine learning algorithm
underlying any of our approaches would need to be tuned to get the best
performance. In this paper, we do not do any tuning because our focus is not on
the absolute performance, but on the performance differences between the
approaches with everything else being equal.

The results shown in the figures are not immediately conclusive with respect to
which rank prediction approaches provide the best performance. We present the
results in aggregate form in tables~\ref{tab:sumsbest} and~\ref{tab:sumsworst}.

\begin{table*}[htb]
\centering
\begin{tabular}{rlllll}
\toprule
& CSP & QBF & SAT-HAN & SAT-IND & total\\
\midrule
Order & 3 & 2.4 & 2.789 & 3.148 & 11.337\\
Order score (classification) & 3 & 1.5 & 2.625 & 2.527 & 9.652\\
Order score (regression) & 3 & 1.2 & 2.777 & 2.871 & 9.848\\
Faster than classification & 3 & 2.3 & 2.907 & 3.29 & 11.497\\
Faster than difference classification & 3 & 1.5 & 2.626 & 2.909 & 10.036\\
Solve time & 3 & -0.45 & 1.563 & 1.379 & 5.492\\
Solve time (log) & 3 & 0.2 & 2.453 & 2.761 & 8.413\\
Probability of being best & 3 & 0.3 & 0.496 & 1.25 & 5.046\\
Faster than majority vote & 3 & 1.9 & 2.711 & 3.052 & 10.662\\
Faster than difference sum & 3 & 0.2 & 1.544 & 1.594 & 6.338\\
\bottomrule
\end{tabular}
\caption{Sum of the ranking quality scores over all quartiles for all data sets
and rank prediction approaches for the best machine learning algorithm for a
particular prediction approach. Higher scores are better. All numbers are
rounded to three decimal places.\label{tab:sumsbest}}
\end{table*}

\begin{table*}[htb]
\centering
\begin{tabular}{rlllll}
\toprule
& CSP & QBF & SAT-HAN & SAT-IND & total\\
\midrule
Order & 3 & 1 & 2.584 & 2.821 & 9.406\\
Order score (classification) & 3 & 0.8 & 1.705 & 1.357 & 6.862\\
Order score (regression) & 3 & 1.05 & 2.672 & 2.678 & 9.4\\
Faster than classification & 3 & 1 & 1.981 & 2.741 & 8.722\\
Faster than difference classification & 1 & -0.5 & 1.246 & 1.454 & 3.2\\
Solve time & 1 & -0.8 & 1.667 & 1.409 & 3.275\\
Solve time (log) & -3 & 0.5 & 2.442 & 2.724 & 2.667\\
Probability of being best & 3 & -0.8 & 0.267 & 0.068 & 2.535\\
Faster than majority vote & 3 & 1.2 & 1.226 & 1.44 & 6.866\\
Faster than difference sum & 1 & -0.8 & 1.667 & 1.409 & 3.275\\
\bottomrule
\end{tabular}
\caption{Sum of the ranking quality scores over all quartiles for all data sets
and rank prediction approaches for the worst machine learning algorithm for a
particular prediction approach. Higher scores are better. All numbers are
rounded to three decimal places.\label{tab:sumsworst}}
\end{table*}

The tables show the sum of the quartiles of the Spearman correlation scores for
all data sets and rank prediction approaches. Again we show the numbers for both
the best and the worst machine learning algorithm. We decided to use the sum for
comparison because it takes into account all the scores. In addition to the
scores for the individual data sets, we show the total sum over all data sets.

The results demonstrate that some approaches are more susceptible than others to
the performance of the underlying machine learning algorithms. This becomes
clear from the spread between the best and worst algorithm -- in some cases,
there is almost no difference at all, whereas in other cases there is a
significant difference.

The overall best approach is the Faster than classification approach, closely
followed by the Order approach. The Faster than majority vote and Order score
(regression) approaches exhibit good performance as well. Looking at the results
for the worst machine learning algorithm, the best approach is Order, closely
followed by Order score (regression), Faster than classification and Faster than
majority vote.

The results shown in tables~\ref{tab:sumsbest} and~\ref{tab:sumsworst} are
consistent with respect to which approaches perform well and which ones do not.
In both cases, the worst approach is Probability of being best. This suggests
that there is a difference between the different rank prediction approaches that
is independent of the underlying machine learning algorithm.

We performed the Kruskal-Wallis rank sum test across the different approaches.
The differences between the series of quartiles over all data sets were not
statistically significant. The Wilcoxon signed-rank test for pairs of approaches
did suggest statistically significant differences though. This suggests that
there is not enough data to draw statistically significant conclusions on the
entire set of approaches, but that there are differences.

The results clearly demonstrate that the relationship between the portfolio
algorithms is important to take into account when predicting the ranking of
algorithms. In general, the approaches that consider the algorithms only in
isolation perform worse than the approaches that consider the portfolio as a
whole or pairs of algorithms. The exception to this rule is the Order score
(regression) approach, which consistently performs well.

While this result is intuitively plausible and not unexpected, this paper is the
first to investigate it through rigorous empirical evaluation. While
\cite{xu_hydra-mip_2011} use a similar approach that considers pairs of
algorithms, they do not compare it to other approaches. We show conclusively
that this is beneficial in terms of the similarity of the derived ranking to the
actual ranking.

We were mildly surprised by the good performance of the Order approach. It is
the conceptually simplest of the approaches we evaluated and the rationale for
including it at all was only to provide a comparison to the other approaches.
Nevertheless, it turns out that this simple approach is capable of producing
good rankings even for portfolios with a relatively large number of solvers
where the space of possible labels is large. This performance depends of course
also on the number of rankings that are actually going to be predicted, as a
potentially large subset of all rankings will never occur in the training data.
With this caveat in mind, the Order approach can provide a reasonable starting
point.

\subsection{Performance in $\mathcal{R}$}

Overall, the approaches in $\mathcal{R}_1$ perform better than those in
$\mathcal{R}_2$. Out of five approaches at each level, only one in
$\mathcal{R}_2$ features in the overall top four approaches. Level
$\mathcal{R}_2$ also contains the overall worst approach. This is especially
apparent in tables~\ref{tab:sumsbest} and~\ref{tab:sumsworst}, where the top
half of the table contains the approaches from $\mathcal{R}_1$ and the bottom
half the ones from $\mathcal{R}_2$. In both tables, the top half is consistently
better than the bottom half.

The reason for this is that the predictions approaches in $\mathcal{R}_2$ have
to make are inherently more complex and there is more margin for error.
Statistically, approaches that make simple predictions have a better chance of
being correct simply by luck. While a machine learning approach would ideally be
able to identify and exploit the relationship between problem features and the
ranking of the portfolio algorithms, all statistical approaches such as the ones
used here benefit from having a higher chance of being correct by pure luck.

It should be noted that the only criteria we use for evaluating the quality of
an approach is the computed ranking itself. The approaches in $\mathcal{R}_2$
provide more information than that and, more importantly, are evaluated
according to a different objective criteria during training. However, the two
different objectives are closely linked. While our evaluation puts the
approaches in $\mathcal{R}_2$ at a slight disadvantage, they still demonstrate
good performance overall.

In the machine learning literature, great care is usually taken to train and
evaluate according to the same objective criteria. We consciously do not
restrict ourselves to the approaches in $\mathcal{R}_1$ to facilitate this, but
include other approaches as well to give a better picture of the performance of
a wider variety of methods. While this may seem strange to machine learning
researchers, it is common practice in algorithm selection -- the original
SATzilla for example trained models to predict the performance of algorithms,
but was evaluated in terms of how often the algorithm with the best predicted
performance was the one with the actual best performance.

\clearpage

\section{Conclusions and future work}

We have presented and evaluated several approaches for predicting the ranking of
algorithms from a portfolio on a particular problem. While there are a vast
number of publications that propose and evaluate methods for choosing the best
algorithm, few are concerned with predicting the complete ranking of the
algorithms. This information is becoming increasingly relevant in many
applications however, for example through the increase of the number of
processors and thus the potential of running several algorithms at the same
time.

Many approaches reported in the literature rely at least implicitly on rankings
of algorithms, for example by computing schedules according to which to run the
algorithms in a portfolio. Despite this, no study of how to make this prediction
in practice has been presented so far. It is this gap in the literature that we
have addressed in this paper.

In addition to the empirical evaluation of a range of approaches, we presented a
framework to assess the power of predictions with respect to deriving a ranking.
This allows us to classify the approaches that we have evaluated and compare
their predictive output. Our framework was inspired by the theory of scales of
measurement.

One of the main and most important conclusions of this paper is that rank
prediction approaches that consider algorithms in isolation perform worse than
approaches that consider them in combination. This is not surprising, given that
a ranking is intrinsically concerned with the relationship of algorithms.

We identified the Faster than classification, Faster than majority vote and
Order as the approaches that deliver the best overall performance. While some of
these are complex and rely on layers of machine learning models, the Order
approach is actually the simplest of those evaluated here. Its simplicity makes
it easy to implement and an ideal starting point for researchers planning to
predict rankings of algorithms. In addition to the approaches named above,
predicting the order through a ranking score predicted by a regression algorithm
achieved good performance.

\medskip

In our proposed framework, most of the approaches that deliver good performance
are in the level that provides simpler predictions. For many applications it is
desirable however to have the additional information approaches at the higher
level provides. Developing more approaches that are able to deliver this and
provide good performance is a possible avenue for future work.

The question of how to predict a ranking of algorithms is only the first step
for putting this approach to practical use. In the future, we would like to
evaluate different practical approaches for using the rankings -- for example
running the top $n$ algorithms in parallel or computing an explicit schedule
based on ranking scores. Using predicted rankings in practice poses additional
challenges because not only the overall quality of the ranking matters, but also
whether the algorithms with the best performance are actually at the top of the
ranking.

Investigating this question is just one of the many possible avenues for future
work. We believe that with the current explosion of the number of processors
that are available even in consumer-grade machines and thus the increased
ability to run more and more algorithms in parallel, the ability to reliably
predict good rankings will be increasingly important for practical applications.

\subsubsection*{Acknowledgements}

Lars Kotthoff is supported by European Union FP7 grant 284715.

\bibliography{\jobname}
\bibliographystyle{splncs03}

\end{document}